\documentclass{article}

\usepackage{microtype}
\usepackage{graphicx}
\usepackage{subfigure}
\usepackage{booktabs} 
\usepackage{multirow}
\usepackage{comment}

\usepackage{hyperref}

\usepackage[accepted]{icml2025}

\usepackage{amsmath}
\usepackage{amssymb}
\usepackage{mathtools}
\usepackage{amsthm}
\usepackage{dsfont}
\usepackage{xcolor}

\usepackage[capitalize,noabbrev]{cleveref}

\theoremstyle{plain}

\theoremstyle{definition}

\theoremstyle{remark}

\usepackage[textsize=tiny]{todonotes}

\icmltitlerunning{Submission and Formatting Instructions for ICML 2025}

\begin{document}

\twocolumn[
\icmltitle{Classifier Reconstruction Through Counterfactual-Aware Wasserstein Prototypes}



\icmlsetsymbol{equal}{*}

\begin{icmlauthorlist}
\icmlauthor{Xuan Zhao}{yyy}
\icmlauthor{Zhuo Cao}{yyy}
\icmlauthor{Arya Bangun}{yyy}
\icmlauthor{Hanno Scharr}{yyy}
\icmlauthor{Ira Assent}{yyy,sch}
\end{icmlauthorlist}

\icmlaffiliation{yyy}{IAS-8, Forschungszentrum Jülich, Germany}
\icmlaffiliation{sch}{Department of Computer Science, Aarhus University, Denmark}

\icmlcorrespondingauthor{Xuan Zhao}{xu.zhao@fz-juelich.de}

\icmlkeywords{Machine Learning, ICML}

\vskip 0.3in
]



\printAffiliationsAndNotice{\icmlEqualContribution} 

\begin{abstract}
Counterfactual explanations provide actionable insights by identifying minimal input changes required to achieve a desired model prediction. Beyond their interpretability benefits, counterfactuals can also be leveraged for model reconstruction, where a surrogate model is trained to replicate the behavior of a target model. In this work, we demonstrate that model reconstruction can be significantly improved by recognizing that counterfactuals, which typically lie close to the decision boundary, can serve as informative—though less representative—samples for both classes. This is particularly beneficial in settings with limited access to labeled data. We propose a method that integrates original data samples with counterfactuals to approximate class prototypes using the Wasserstein barycenter, thereby preserving the underlying distributional structure of each class. This approach enhances the quality of the surrogate model and mitigates the issue of decision boundary shift, which commonly arises when counterfactuals are naively treated as ordinary training instances. Empirical results across multiple datasets show that our method improves fidelity between the surrogate and target models, validating its effectiveness. 

\end{abstract}

\section{Introduction}
\label{intro}

Counterfactual explanations have gained prominence as a growing research field \cite{DBLP:journals/corr/abs-1711-00399,DBLP:journals/datamine/Guidotti24,DBLP:conf/fat/BarocasSR20,DBLP:journals/csur/VermaBHHDS24} for offering actionable insights into altering model outcomes—for instance, suggesting a \$10K income increase to secure a loan. Notably, counterfactuals can inadvertently expose model internals, creating a delicate balance between transparency and privacy \cite{DBLP:journals/corr/abs-2009-01884, DBLP:conf/fat/WangQM22, DBLP:conf/nips/DissanayakeD24}. This poses risks in Machine Learning as a Service platforms, where adversaries might exploit counterfactual queries to replicate models via surrogate training, a tactic known as model extraction \cite{DBLP:journals/cm/GongWCYJ20}. Companies must therefore assess information leakage when deploying counterfactuals. Conversely, model reconstruction can empower applicants to gauge approval odds without formally submitting sensitive data—crucial in scenarios with limited attempts (e.g., credit applications affecting scores). This work formalizes the fidelity of model reconstruction from counterfactual queries.

Prior methods treat counterfactuals as labeled data to train surrogate models \cite{DBLP:journals/corr/abs-2009-01884}. While effective for balanced queries near the decision boundary, imbalanced datasets induce decision boundary shifts, where surrogate boundaries diverge from the target model’s \cite{DBLP:conf/nips/DissanayakeD24}. This stems from margin-based generalization \cite{DBLP:conf/aies/ShokriSZ21} and worsens with one-sided counterfactuals (e.g., only reject-to-accept modifications). Two-sided queries could mitigate this \cite{DBLP:conf/fat/WangQM22}, but such strategies fail when only one-sided counterfactuals are available—a common yet challenging scenario (e.g., loan rejections lacking accept-to-reject examples). Counterfactual Clamping loss function \cite{DBLP:conf/nips/DissanayakeD24} offers an approach to reconstruct the classifier using neural networks by adjusting the classic entropy loss. However, when applied to scenarios with limited query samples, there's a heightened risk of overfitting. 

We propose a surrogate model based on prototype-driven classification to address the challenge of limited access to queries from the original classifier and its counterfactuals. Counterfactuals are assigned a soft label of 0.5, which helps address class imbalance while simultaneously mitigating issues arising from limited sample sizes. We adopt an approach by leveraging Wasserstein barycenters to represent soft prototypes for each class. This method captures the underlying structure of each class distribution in a data-efficient manner, making it particularly suitable for low-data regimes.

\section{Related Work}

\subsection{Counterfactual Explanations in Classification}
Counterfactual explanations, first formalized by \citet{DBLP:journals/corr/abs-1711-00399}, have become a cornerstone of explainable AI. Counterfactual generation is typically formalized as a constrained optimization problem. Given a model \( f \), an input \( x \), and a desired prediction \( y' \), the goal is to find a counterfactual \( x' \) such that:
\[
\min_{x'} \; \mathcal{L}(f(x'), y') + \lambda \cdot d(x, x')
\]
where \( \mathcal{L} \) is a loss function encouraging the prediction \( f(x') \) to match \( y' \), \( d(x, x') \) is a distance metric, and \( \lambda \) is a regularization parameter. This framework is widely used across methods for producing counterfactuals that are both faithful and interpretable.

Notably, \citet{vanlooveren2021interpretable} develop prototype-based counterfactuals using standard distance metrics. Our methodology, however, integrates counterfactual constraints directly into the barycenter optimization process.

\subsection{Prototype-Based Classification}
Prototype-based classification is a foundational approach in machine learning, where each class is represented by one or more prototypes—representative examples that encapsulate the characteristics of the class. Classification decisions are made by comparing new instances to these prototypes using a suitable distance metric. \citet{biehl2016prototype} provide a comprehensive overview of prototype-based models, discussing extensions like relevance learning, which adapts the distance metric to emphasize relevant features for classification. In the context of few-shot learning, where the goal is to classify instances with limited labeled examples, Prototypical Networks have gained prominence. \citet{snell2017prototypical} propose learning a metric space where classification is performed by computing distances to class prototypes, defined as the mean of embedded support examples. Prototype-based methods are particularly advantageous in scenarios with limited data. By summarizing each class with representative prototypes, these methods can generalize effectively even when only a few labeled examples are available.

\subsection{Wasserstein Barycenters in Machine Learning}

The Wasserstein distance, rooted in optimal transport theory, provides a meaningful metric for comparing probability distributions by considering the geometry of the underlying space.

The \emph{2-Wasserstein distance} \( W_2 \) between two probability measures \( \mu \) and \( \nu \) on a metric space \( (M, d) \) is defined as:
\[
W_2^2(\mu, \nu) = \inf_{\gamma \in \Gamma(\mu, \nu)} \int_{M \times M} d(x, y)^2 \, d\gamma(x, y)
\]
Here: \( \Gamma(\mu, \nu) \) denotes the set of all couplings (i.e., joint distributions) with marginals \( \mu \) and \( \nu \); \( d(x, y) \) is the distance between points \( x \) and \( y \) in the space \( M \).

Wasserstein barycenters extend this concept by offering a method to compute the ``average" of multiple probability distributions, preserving the intrinsic geometric relationships among them. Such barycenters have been effectively utilized as prototypes in various machine learning applications.

Given a set of probability measures \( \mu_1, \mu_2, \ldots, \mu_N \) defined on a metric space \( (M, d) \), and associated non-negative weights \( \lambda_1, \lambda_2, \ldots, \lambda_N \) summing to 1, the \emph{2-Wasserstein barycenter} \( \nu^* \) is defined as the probability measure that minimizes the weighted sum of squared 2-Wasserstein distances to the given measures:
\[
\nu^* = \arg\min_{\nu \in \mathcal{P}_2(M)} \sum_{i=1}^N \lambda_i W_2^2(\nu, \mu_i)
\]
Here: \( \mathcal{P}_2(M) \) denotes the set of probability measures on \( M \) with finite second moments. \( W_2(\nu, \mu_i) \) represents the 2-Wasserstein distance between \( \nu \) and \( \mu_i \).

In the context of prototype-based classification, utilizing Wasserstein barycenters allows for the aggregation of class-specific data distributions into a single representative prototype. This approach is particularly beneficial in scenarios with limited data, as it provides a robust summary that accounts for the variability within each class. Furthermore, the Wasserstein barycenter's sensitivity to the underlying geometry of the data ensures that the prototypes maintain the structural relationships present in the original distributions. \citet{DBLP:conf/cvpr/ZenR11} use Earth Mover’s Distance based prototypes which are actually Wasserstein barycenters.

\section{Method}
\label{method}

\subsection{Problem Setting}

We examine a binary classifier \( m: \mathbb{R}^d \rightarrow [0,1] \) that processes input vectors \( \boldsymbol{x} \in \mathbb{R}^d \) and generates probability scores. The final classification decision $\hat{y}$ is made by applying a 0.5 threshold to the model's output probability: $\hat{y}_m = \mathbb{I}_{[0,1]}[m(\boldsymbol{x}) \geq 0.5]$,
where \( \mathbb{I}_{[0,1]}[\cdot] \) is the binary indicator function. The model is accompanied by a counterfactual generator \( g_m \) that creates perturbed examples lying close to the classification boundary. In our framework, we assume access to a pre-trained target model \( m \) that responds to user queries with prediction outputs. The counterfactual mechanism \( g_m \) is specifically activated for cases where \( m \) predicts class 0 (i.e., when \( m(\boldsymbol{x}) < 0.5 \)). The primary objective is to learn an approximation model \( \hat{m} \) that replicates the target model's behavior while minimizing the number of necessary queries.
To reconstruct the model \(m\), we reformulate the problem over a feature space \(\mathcal{X} \subseteq \mathbb{R}^d\) and a structured label space \(\mathcal{Y} = \{0, 0.5, 1\}\). Here, label 0 corresponds to samples from class 0, label 1 to samples from class 1, and label 0.5 represents counterfactual or ambiguous examples that blend characteristics of both classes. These counterfactuals are assumed to lie near the boundary and are treated as soft samples of both classes.
The dataset for class 0 is denoted by \(\mathcal{D}_0 = \{(x_i^{(0)}, 0)\}_{i=1}^{N_0}\), where \(x_i^{(0)} \sim \mathbb{P}_0\). Similarly, the dataset for class 1 is \(\mathcal{D}_1 = \{(x_i^{(1)}, 1)\}_{i=1}^{N_1}\), with \(x_i^{(1)} \sim \mathbb{P}_1\). The counterfactual dataset is given by \(\mathcal{D}_{cf} = \{(x_i^{(cf)}, 0.5)\}_{i=1}^{N_{cf}}\), where \(x_i^{(cf)} \sim \mathbb{P}_{cf}\) represents samples that mix features from both classes.

\subsection{Model Reconstruction using Wasserstein Barycenters as Prototypes}\label{method _bary}

To represent each class in a way that incorporates both original and counterfactual samples, we compute a soft prototype for each class using the Wasserstein barycenter. For every class \(c \in \{0, 1\}\), we define a barycenter distribution \(\mathbb{Q}_c\) as the solution to the following optimization problem:
\begin{equation}
    \mathbb{Q}_c = \arg \min_{\mathbb{Q} \in \mathbb{P}(\mathcal{X})} \left( W_2^2(\mathbb{Q}, \mathbb{P}_c) + \lambda_c W_2^2(\mathbb{Q}, \mathbb{P}_{cf}) \right)
\end{equation}
Here, \(W_2^2(\cdot, \cdot)\) denotes the squared 2-Wasserstein distance between probability distributions, which measures the optimal transport cost of transforming one distribution into another. Intuitively, it captures both the amount of mass that needs to be moved and the distance over which it must be moved, making it a meaningful geometric measure of similarity between distributions. The coefficient \(\lambda_c = 0.5\) balances the influence of the original class distribution \(\mathbb{P}_c\) and the counterfactual distribution \(\mathbb{P}_{cf}\) in determining the barycenter \(\mathbb{Q}_c\).

To ensure that the counterfactual samples are symmetrically situated between the class prototypes, we introduce a regularization term:
\begin{equation}
    \mathcal{R}(\mathbb{Q}_0, \mathbb{Q}_1) = \left( W_2(\mathbb{Q}_0, \mathbb{P}_{cf}) - W_2(\mathbb{Q}_1, \mathbb{P}_{cf}) \right)^2
\end{equation}
This term penalizes asymmetry in the distances from the counterfactual distribution to the class barycenters, encouraging the counterfactuals to be equidistant from both sides of the decision boundary.

The overall objective function combines the class-wise barycenter losses with the symmetry regularization:
\begin{equation}
    \min_{\mathbb{Q}_0, \mathbb{Q}_1} \sum_{c \in \{0,1\}} \left( W_2^2(\mathbb{Q}_c, \mathbb{P}_c) + \lambda_c W_2^2(\mathbb{Q}_c, \mathbb{P}_{cf}) \right) + \gamma \mathcal{R}(\mathbb{Q}_0, \mathbb{Q}_1)
\end{equation}
Here, \(\gamma > 0\) is a regularization coefficient that controls the strength of the symmetry constraint. The result is a pair of barycentric prototypes that not only reflect their respective class distributions but also respect the structure of the decision boundary as implied by the counterfactuals.

Given the learned barycenters \(\mathbb{Q}_0\) and \(\mathbb{Q}_1\), we define a classification rule for assigning labels to new inputs. For a test sample \(x \in \mathcal{X}\), we compare its distance to the two barycenters using the Wasserstein-2 metric. Let \(\delta_x\) denote the Dirac delta distribution centered at \(x\). Then, the predicted label \(\hat{y}(x)\) is given by:
\begin{equation}
    \hat{y}_{\hat{m}}(x) = 
    \begin{cases}
        0 & \text{if } W_2(\delta_x, \mathbb{Q}_0) < W_2(\delta_x, \mathbb{Q}_1) - \tau \\
        1 & \text{if } W_2(\delta_x, \mathbb{Q}_1) < W_2(\delta_x, \mathbb{Q}_0) - \tau 
    \end{cases}
\end{equation}
The threshold parameter \(\tau \geq 0\) introduces a margin to avoid overly confident predictions near the decision boundary. 

We employ fidelity \cite{DBLP:journals/corr/abs-2009-01884} as a metric to evaluate the performance of model reconstruction. Given a target model \( m \) and a reference dataset \(\mathcal{D}_{\text{ref}}\),  the fidelity of a surrogate model \(\hat{m}\) is defined as

%
\[
\text{Fid}_{m,\mathcal{D}_{\text{ref}}}(\hat{m}) = \frac{1}{|\mathcal{D}_{\text{ref}}|} \sum_{\boldsymbol{x} \in \mathcal{D}_{\text{ref}}} \mathbb{I}_{[0,1]} \left[\hat{y}_m(\boldsymbol{x})= \hat{y}_{\hat{m}}(\boldsymbol{x}) \right].
\]

\section{Experiment}
\begin{table*}[h]
\centering
\caption{Average fidelity with $500$, $400$, $300$ queries on the datasets}
\resizebox{1.0\linewidth}{!}{%
\begin{tabular}{lcccccccccc}
\toprule
 &\multicolumn{3}{c}{Query Size (500)} & \multicolumn{3}{c}{Query Size (400)} & \multicolumn{3}{c}{Query Size (300)} \\
\cmidrule(lr){2-4} \cmidrule(lr){5-7}\cmidrule(lr){8-10}
 & Baseline 1 & Baseline 2 & Ours & Baseline 1 & Baseline 2 & Ours & Baseline 1 & Baseline 2 & Ours \\
\midrule
Adult In. & $91 \pm 3.2$ & $94 \pm 3.2$ & $96 \pm 2.5$ & $89 \pm 3.5$ & $92 \pm 3.5$ & $94 \pm 2.8$ & $87 \pm3.8$ & $90 \pm 3.8$ & $93 \pm 3.2$ \\
COMPAS & $92 \pm 3.2$ & $94 \pm 2.0$ & $96 \pm 2.3$ & $90 \pm 3.5$ & $92 \pm 2.3$ & $94 \pm 2.6$ & $88 \pm 3.8$ & $90 \pm 2.6$ & $94 \pm 3.0$ \\
DCCC & $89 \pm 8.9$ & $91 \pm 0.9$ & $97 \pm 1.5$ & $87 \pm 9.2$ & $89 \pm 1.2$ & $95 \pm 1.8$ & $85 \pm 9.5$ & $87 \pm 1.5$ & $93 \pm 2.1$ \\
HELOC & $91 \pm 4.7$ & $93 \pm 2.2$ & $95 \pm 2.0$ & $89 \pm 5.0$ & $91 \pm 2.5$ & $93 \pm 2.3$ & $87 \pm 5.3$ & $89 \pm 2.8$ & $93 \pm 2.6$ \\
\bottomrule
\end{tabular}}
\end{table*}

We conduct a series of experiments to evaluate the effectiveness of our proposed method for reconstructing binary classifiers using Wasserstein barycenters as class prototypes. We update our Wasserstein barycenters with the method \textit{Barycenters with free support} in python package POT \cite{flamary2021pot}. We compare our approach against two SOTA methods: the first is an attack technique for classifier reconstruction without primer knowledge to training data, as described in \citet{DBLP:journals/corr/abs-2009-01884}, referred to as \textit{Baseline 1} where counterfactuals are treated as normal samples; the second is a method that modifies the standard entropy loss to incorporate counterfactual explanations, proposed by \citet{DBLP:conf/nips/DissanayakeD24}, referred to as \textit{Baseline 2}. The target classifiers are trained using logistic regression on a training dataset \(\mathcal{D}_{\text{train}}\), which remains unknown during the reconstruction phase. To generate initial one-sided counterfactuals, we adopt the Minimum Cost Counterfactuals (MCCF) method \cite{DBLP:journals/corr/abs-1711-00399}. We evaluate reconstruction performance using the \textit{fidelity} metric described in Section \ref{method _bary}, which measures the agreement between the predictions of the target model and the reconstructed (surrogate) model. Fidelity is assessed over a set of test samples drawn from the original data manifold, and used as the reference dataset \(\mathcal{D}_{\text{ref}}\). Experiments are conducted on four datasets: Adult Income, COMPAS, DCCC, and HELOC (see Appendix for details). Table~\ref{tab_average_fidelity} summarizes the \textit{fidelity} results across these datasets. In all cases, our method achieves either superior or comparable fidelity relative to Baseline 1 and Baseline 2. Notably, our approach demonstrates a clear advantage when the query size is small, highlighting its efficiency in low-query regimes.

\paragraph{Other Architectures for Reconstruction:} 

We also compare our method to Baseline 2 using neural network surrogate models of varying complexity, as shown in Table \ref{table:complex} in the Appendix. As the neural network architecture in Baseline 2 becomes more complex, our method consistently maintains strong performance, whereas Baseline 2 exhibits noticeable degradation. This is particularly striking given that more complex neural networks theoretically have greater capacity to approximate arbitrary functions. The decline in Baseline 2’s performance is especially evident under limited query budgets, where its reliance on training a neural network surrogate leads to overfitting due to insufficient data.

\paragraph{Other Counterfactual Generation Techniques:} We investigate the impact of various counterfactual generation methods, focusing on attributes such as sparsity, actionability, realism, and robustness. To generate sparse counterfactuals, we employ an $L_1$-norm as the cost function, promoting minimal feature changes. For actionable counterfactuals, we utilize the DiCE framework \cite{DBLP:conf/fat/MothilalST20}, which allows the specification of immutable features to ensure feasibility. Realistic counterfactuals, those that lie close to the data manifold, are produced by identifying the nearest neighbor from the desired class and by applying the C-CHVAE method \cite{DBLP:conf/www/PawelczykBK20}, which leverages variational autoencoders. To enhance robustness, we generate counterfactuals using the ROAR approach \cite{DBLP:conf/nips/UpadhyayJL21}, designed to maintain validity under model shifts. We assess the effectiveness of these methods through attack performance evaluations on the Adult dataset (Table~\ref{table:ce}).
The quality of counterfactual examples significantly influences the fidelity of model reconstruction. High-quality counterfactuals—those that are plausible and lie close to the data manifold enable surrogate models to more accurately approximate the decision boundaries of the original model. C-CHVAE does not perform well, which we attribute to the generally weaker performance of generative models on tabular datasets.

\begin{table}[h]
\caption{Fidelity with different counterfactual methods on Adult dataset}
\label{table:ce}
\centering
\scalebox{0.68}{
\begin{tabular}{lccc}
\toprule
 & \textbf{Query Size (500)} & \textbf{Query Size (400)} & \textbf{Query Size (300)} \\
\midrule
MCCF & $96 $ & $94 $ & $93 $ \\
DiCE  & $96 $ & $94 $ & $92 $ \\
1-Nearest-Neightbor & $97 $ & $91 $ & $89 $ \\
ROAR  & $95 $ & $93 $ & $91$ \\
C-CHVAE   & $83 $ & $79$ & $77 $ \\
\bottomrule
\end{tabular}
}
\end{table}

\section{Discussion and Future Work}

This work demonstrates that Wasserstein barycenters provide a robust framework for classifier reconstruction, particularly in scenarios with limited data and the presence of counterfactual examples. By integrating optimal transport theory, our approach captures nuanced relationships between data distributions, enabling the formation of soft prototypes that effectively represent both labeled data and counterfactuals. This representation-centric perspective is particularly beneficial when dealing with small or noisy samples, where traditional methods may falter. Analyses reveal that our method maintains high fidelity in model reconstruction. Notably, in small data regimes, where overfitting and poor generalization are prevalent concerns, our approach exhibits superior stability and flexibility compared to baseline methods. This performance underscores the significance of incorporating geometric and distributional considerations into the reconstruction process.

Future work should aim to quantitatively investigate the impact of sample size on the fidelity of model reconstruction. Additionally, it would be valuable to explore alternative representations of the dataset—for instance, using prototype representations other than the barycenter—to potentially improve performance. Our current approach assumes no prior knowledge of the original model; however, incorporating such prior information could substantially enhance reconstruction quality. Advancing research in these directions will contribute to developing more data-efficient and interpretable model extraction techniques for real-world applications.

\bibliography{bib}
\bibliographystyle{icml2025}

\newpage
\appendix
\onecolumn
\section{Experimental Setup and Supplementary Findings}

\subsection{Description of Real-World Benchmark Datasets}

To assess the fedelity, we employed four publicly available tabular datasets: Adult Income, COMPAS, DCCC, and HELOC. Below are their key characteristics:

Adult Income \cite{adult}: Derived from the 1994 U.S. Census, this dataset captures demographic and financial attributes such as education level, marital status, age, and annual earnings. The classification task involves predicting whether an individual’s income exceeds \$50,000 (denoted as $y=1$). The original dataset consists of 32,561 entries, with 24,720 labeled as $y=0$ and 7,841 as $y=1$. To balance the classes, we randomly selected 7,841 samples from $y=0$, resulting in a final dataset of 15,682 entries. The dataset includes 6 numerical and 8 categorical features, with the latter converted to integer encodings. All features were normalized to $[0,1]$.

Home Equity Line of Credit (HELOC): This dataset records credit risk assessments for customers seeking home equity loans \cite{fico2018}. It comprises 10,459 entries, each with 23 numerical features. The prediction target, ``is\_at\_risk," identifies customers likely to default. The dataset is moderately imbalanced, with 5,000 samples for $y=0$ and 5,459 for $y=1$. For our experiments, we used a subset of 10 key features, including ``estimate\_of\_risk," ``net\_fraction\_of\_revolving\_burden," and ``percentage\_of\_legal\_trades," all scaled to $[0,1]$.

COMPAS: Developed to study racial bias in recidivism prediction algorithms, this dataset contains 6,172 entries with 20 numerical features \cite{propublica2016compas}. The target variable, ``is\_recid," divides the data into 3,182 ($y=0$) and 2,990 ($y=1$) samples. Feature values were normalized to $[0,1]$.

Default of Credit Card Clients (DCCC): This dataset tracks credit card payment behaviors in Taiwan \cite{yeh2009default}, with the goal of predicting defaults (``default.payment.next.month"). The original dataset has 30,000 entries (23,364 for $y=0$, 6,636 for $y=1$). To address class imbalance, we randomly subsampled 6,636 instances from $y=0$. Categorical features were integer-encoded, and all attributes were normalized to [0,1].

\subsection{Experiment Details}

All experiments were conducted on a machine equipped with an NVIDIA RTX 3090 GPU. The regularization coefficient $\gamma$  is set to be 0.3.

\begin{table*}[h]
\centering
\caption{Average fidelity with 500, 400, 300 queries on the datasets. Baseline 2.1 has hidden layers with neurons (20, 10, 5) and Baseline 2.2 has hidden layers with neurons (20, 10) }
\label{table:complex}
\resizebox{1.0\linewidth}{!}{%
\begin{tabular}{lcccccccccc}
\toprule
 &\multicolumn{3}{c}{Query Size (500)} & \multicolumn{3}{c}{Query Size (400)} & \multicolumn{3}{c}{Query Size (300)} & \\
\cmidrule(lr){2-4} \cmidrule(lr){5-7}\cmidrule(lr){8-10}
 & Baseline 2.1 & Baseline 2.2 & Ours & Baseline 2.1 & Baseline 2.2 & Ours & Baseline 2.1 & Baseline 2.2 & Ours\\
\midrule
Adult In. & $92 \pm 4.1$ & $94 \pm 3.2$ & $96 \pm 2.5$ & $87 \pm 3.9$ & $92 \pm 3.5$ & $94 \pm 2.8$ & $80 \pm 3.8$ & $90 \pm 3.8$ & $93 \pm 3.2$ \\
COMPAS & $90 \pm 3.0$ & $94 \pm 2.0$ & $96 \pm 2.3$ & $86 \pm 3.9$ & $92 \pm 2.3$ & $94 \pm 2.6$ & $82 \pm 5.8$ & $90 \pm 2.6$ & $94 \pm 3.0$ \\
DCCC & $88 \pm 8.3$ & $91 \pm 0.9$ & $97 \pm 1.5$ & $84 \pm 7.2$ & $89 \pm 1.2$ & $95 \pm 1.8$ & $81 \pm 5.5$ & $87 \pm 1.5$ & $93 \pm 2.1$ \\
HELOC & $89 \pm 4.5$ & $93 \pm 2.2$ & $95 \pm 2.0$ & $87 \pm 3.1$ & $91 \pm 2.5$ & $93 \pm 2.3$ & $82 \pm 3.3$ & $89 \pm 2.8$ & $93 \pm 2.6$ \\
\bottomrule
\end{tabular}
}
\end{table*}


\end{document}